# Integer-Only Operations on Extreme Learning Machine Test Time Classification

**Emerson Lopes Machado[a*], Cristiano Jacques Miosso[b], Ricardo Pezzuol Jacobi[a]**
[a]*Dept. of Computer Science, University of Brasilia, Brazil*
[b]*University of Brasilia at Gama, Brazil*
E-mail: *bmi.machado@gmail.com

**Abstract**

We present a theoretical analysis and empirical evaluations of a novel set of techniques for computational cost reduction of test time operations of network classifiers based on extreme learning machine (ELM). By exploring some characteristics we derived from these models, we show that the classification at test time can be performed using solely integer operations without compromising the classification accuracy. Our contributions are as follows: (i) We show empirical evidence that the input weights values can be drawn from the ternary set $\{-1, 0, -1\}$ with limited reduction of the classification accuracy. This has the computational advantage of dismissing multiplications; (ii) We prove the classification accuracy of normalized and non-normalized test signals are the same; (iii) We show how to create an integer version of the output weights that results in a limited reduction of the classification accuracy. We tested our techniques on 5 computer vision datasets commonly used in the literature and the results indicate that our techniques can allow the reduction of the computational cost of the operations necessary for the classification at test time in FPGAs. This is important in embedded applications, where power consumption is limited, and crucial in data centers of large corporations, where power consumption is expensive.

## 1. Introduction

Although most signals nowadays are acquired using an analog-to-digital converter (ADC), which digitalizes and represents them with integer values, virtually all classification algorithms use floating-point operations in the classification process at test time. These floating-point operations have a high cost in hardware, specially in FPGA, as they require a much larger area when compared to integer operations, consuming also more energy and time to operate. Compared to a GPU, an FPGA requires 12 times more density when floating-point operations are necessary [2]. One common solution is to use fixed-point, which trades the manufacturing cost with a decrease in the numerical precision caused by quantization error. This issue is commonly solved by increasing the precision of the accumulator at a higher cost of the DSP and a higher energy consumption [16].

In this paper, we propose a set of techniques that allow to use solely integer operations for the classification at test time of classifiers based on matrix-vector multiplications such as network based classifiers. Our proposed techniques explore some properties we derive regarding these classifiers and how they behave when classifying integer signals.

As a case study for our techniques, we use the classification algorithm Extreme Learning Machine (ELM) [7]. Our tests indicate that our techniques can reduce the computational cost by using only integer operations with a limited loss of classification accuracy. This has a valuable application in embedded systems where power consumption is critical and computational power is restricted. Furthermore, these techniques may dismiss the necessity of using DSPs for intense matrix-vector operations in FPGAs architectures in the context of image classification, lowering the overall manufacturing cost of embedded system.

In this work, all simulations we ran to test our techniques were performed on image classification using the algorithm ELM. Nevertheless, our proposed techniques are sufficiently general to be applied on different problems and different classification algorithms that use matrix-vector multiplications to extract features, such as dictionary based classifiers [4, 15, 13] and Deep Neural Networks (DNN) [14, 17, 8].

## 2. Overview of Extreme Learning Machine

### 2.1. Single Hidden Layer Feedforward Network

Single hidden layer feedforward neural network (SLFN) is a type of neural network that is broadly used in many fields because of its approximation ability in non-linear mapping. Their parameters such as input weights and biases are usually trained using gradient-descent based learning algorithms.

Let $N$ be the number of instances $(\mathbf{x}_j, t_j)$, with $j = 1, \boxed{?}, N$, from the training set $\mathbf{X}$ with labels $\mathbf{T}$, where $\mathbf{x}_j = [x_{j1}, x_{j2}, \boxed{?}, x_{jn}]^T \in \mathbf{R}^n$ are the signal to be modeled and $\mathbf{t}_j = [t_{j1}, t_{j2}, \boxed{?}, t_{jn}]^T \in \mathbf{R}^m$ are their respective labels. A SFLN with $L$ hidden neurons and activation function $g(\cdot)$ that can model $(\mathbf{X}, \mathbf{T})$ may be mathematically modeled as a the linear system

$$\sum_{i=1}^{L} \beta_i g_i(\mathbf{x}_j) = \sum_{i=1}^{L} \beta_i g_i(\mathbf{w}_i^T \mathbf{x}_j + b_i) = \mathbf{o}_j, j = 1, \boxed{?}, N, \qquad (1)$$

where $\mathbf{w}_i = [w_{i1}, w_{i2}, \boxed{?}, w_{in}]^T$ is the input weight vector that connects the input nodes and the $i$ th hidden node, $\beta_i = [\beta_{i1}, \beta_{i2}, \boxed{?}, \beta_{im}]^T$ is the output weight vector that connects the $i$ th hidden node and the output nodes, $b_i$ is the $i$ th hidden node threshold, and $\mathbf{o}_j = [o_{j1}, o_{j2}, \boxed{?}, o_{jn}]^T$ is the output vector of the SFLN.

Using (1) to classify a new instance $\mathbf{x}_j^t$, its predicted class is computed as

$$i^* = \mathrm{argmax}_{i}\{o_i^t\} \ i = 1, 2, \boxed{?}, m,$$

where $m$ is the number of classes of the training set $(\mathbf{x}_j, t_j)$, with $j = 1, \boxed{?}, N$.

A SFLN such as (1) with $L$ hidden nodes can model $(\mathbf{X}, \mathbf{T})$ with zero error, i.e. there exist $g(\cdot)$, $\beta_i$, $\mathbf{w}_i$, and $b_i$ such that

$$\sum_{i=1}^{L}\beta_i g_i(\mathbf{x}_j) = \sum_{i=1}^{L}\beta_i g_i(\mathbf{w}_i^T\mathbf{x}_j + b_i) = \mathbf{t}_j, \ j=1,\ldots,N, \tag{2}$$

which can be compactly represented as

$$\mathbf{H}\beta = \mathbf{T}, \tag{3}$$

where

$$\mathbf{H} = \begin{bmatrix} g(\mathbf{w}_1^T\mathbf{x}_1 + b_1) & \cdots & g(\mathbf{w}_L^T\mathbf{x}_1 + b_L) \\ \vdots & \vdots & \vdots \\ g(\mathbf{w}_1^T\mathbf{x}_N + b_1) & \cdots & g(\mathbf{w}_L^T\mathbf{x}_N + b_L) \end{bmatrix}, \tag{4}$$

$$\beta = \begin{bmatrix} \beta_1^T \\ \vdots \\ \beta_L^T \end{bmatrix}, \text{and } \mathbf{T} = \begin{bmatrix} \mathbf{T}_1^T \\ \vdots \\ \mathbf{T}_N^T \end{bmatrix}. \tag{5}$$

The matrix $\mathbf{H}$ contain the output values of the hidden layer of the SLFN and its $i$th column is the $i$th hidden node output with respect to the instances $\mathbf{X} = [\mathbf{x}_1, \mathbf{x}_2, \ldots, \mathbf{x}_N]$.

Let $\mathbf{x}_{int}$ and $\mathbf{x}$ be respectively a raw vector from the test set and its normalized version, with $\|\mathbf{x}\| = 1$. Also, let $\mathbf{h} = g(\mathbf{W}^T\mathbf{x})$ be the values at the output of the hidden layer of the SFLN, which we call feature vector. Therefore,

$$\begin{aligned} \mathbf{h} &= \max(0, \mathbf{W}^T\mathbf{x}_{int}) \\ &= g(\mathbf{W}^T\mathbf{x}) \\ &= g\left(\mathbf{W}^T \frac{\mathbf{x}_{int}}{\|\mathbf{x}_{int}\|}\right) \\ &= \max\left(0, \mathbf{W}^T \frac{\mathbf{x}_{int}}{\|\mathbf{x}_{int}\|}\right) \\ &= \max\left(0, \frac{1}{\|\mathbf{x}_{int}\|} \mathbf{W}^T\mathbf{x}_{int}\right). \end{aligned}$$

As $\beta$ has been previously trained, the output vector $\mathbf{o}$ is computed as

$$\begin{aligned} \mathbf{o} &= \max\left(0, \frac{1}{\|\mathbf{x}_{int}\|} \mathbf{W}^T\mathbf{x}_{int}\right)\beta \\ &= \max\left(0, \frac{1}{\|\mathbf{x}_{int}\|} \mathbf{W}^T\mathbf{x}_{int}\right)\beta. \end{aligned}$$

In order to train an SFLN such as (1), one shall find specific $g(\cdot)$, $\beta_i$, $\mathbf{w}_i$, and $b_i$ such that minimize the sum of the square of the difference between each expected label vector $\mathbf{o}_j$ and the true label vector $\mathbf{t}_o$, which is the same as

$$\begin{aligned} \underset{\mathbf{H},\beta}{\text{minimize}} \quad & \|\varepsilon\| \\ \text{subject to} \quad & \mathbf{H}\beta - \mathbf{T} - \varepsilon = 0. \end{aligned} \quad (6)$$

Moreover, according to the Bartlett's theory [1], SFLN models that achieves small training errors tend to have better generalization power when their output weights have low energy. Thus, the search for the best SFLN can be represented as the optimization problem

$$\begin{aligned} \underset{\mathbf{H},\beta}{\text{minimize}} \quad & \|\varepsilon\| + \|\beta\| \\ \text{subject to} \quad & \mathbf{H}\beta - \mathbf{T} - \varepsilon = 0, \end{aligned} \quad (7)$$

where $\varepsilon$ is the classification cost function and $\|\beta\|$ is the regularization term of the classification. Traditionally, the search for $\mathbf{H}, \beta$ that satisfy (7) is performed using gradient descent (GD) algorithms, where $g(\cdot)$, $\beta_i$, $\mathbf{w}_i$, and $b_i$ is adjusted iteratively.

Nevertheless, GD algorithms are slow and the optimization iteration may stop in local minima, resulting in low classification accuracy. These are drawbacks that are addressed with Extreme Learning Machine.

## 2.2. Extreme Learning Machine

As aforementioned, gradient-descent based learning algorithms for SFLN may incur in low efficiency because of improper learning steps and may converge to local minima, aside from the high training time. The Extreme Learning Machine (ELM) algorithm [7] were developed to overcome these challenges by using a single-hidden layer feed-forward neural network (SLFN) with the hidden neurons set randomly. The only parameters needed to be tuned are the output weights, which are computed from the output of the hidden layer as follows.

For a SFLN such as (3) with $L = N$ hidden nodes, when its activation function $g(\cdot)$ is infinitely differentiable, the weights and biases can be drawn randomly from any continuous probability distribution. Moreover, the hidden layer output matrix $\mathbf{H}$ is invertible and $\|\mathbf{H}\beta - \mathbf{T}\| = 0$. Also, if the $L \leq N$, this SFLN can approximately model $(\mathbf{X}, \mathbf{T})$ with $\|\mathbf{H}\beta - \mathbf{T}\| < \varepsilon$ [7].

The activation function $g(\cdot)$ can be any infinitely differentiable such as sigmoid functions, radial basis, sine, cosine, exponential, and many other non-regular functions, including the rectifier function [7]. The output weights of the ELM can be analytically determined by the minimum norm least-squares solutions of a general system of linear equations, which can be efficiently computed using a Moore-Penrose pseudoinverse operation. This makes ELM thousand times faster than traditional backpropagation solved using GD algorithms in SFLN.

## 2.3. Regularized Extreme Learning Machine

Even though SFLN such as ELM generalize well, as shown in Section 2.1, ELM is still prone to overfit the model to the training set. Moreover, ELM provides no control over its adaptability power to the training set, since it directly computes the least-squares solution with minimum norm.

It is shown in [3] that ELM can be extended to regularize the built classifier $\beta$ and, thus, increase its generalization power to unseen data. It is based on the principle of structural risk minimization (SRM) from the statistical learning theory [19]. This theory states that the learning real prediction risk is the sum of the empirical risk and the structural risk. In order to have a good generalization power, a model should have the best balance between those risks. The empirical risk is the one computed by the model, which in the ELM case is the loss function $\varepsilon = \|\mathbf{H}\beta - \mathbf{T}\|$. The structural risk is associated with the generated classifier $\beta$ and it is computed to maximize the distance between the closest data and the hyperplane classifier that segregates the classes, which is computed as $\|\beta\|$ [19].

Therefore, the optimization problem (7) is modified to include a regularization factor and becomes

$$\begin{array}{ll} \underset{\mathbf{H},\beta}{\text{minimize}} & \gamma \frac{1}{2}\|\varepsilon\| + \frac{1}{2}\|\beta\| \\ \text{subject to} & \mathbf{H}\beta - \mathbf{T} - \varepsilon = 0, \end{array} \tag{8}$$

where $\gamma$ is the regularization factor and the value $\frac{1}{2}$ that multiplies both terms is just to facilitate the computation of the gradient. The parameter $\gamma$ allows to adjust the trade off between the empirical risk and the structural risk in order to achieve the best generalization power [3]. The Lagrangian of (8) is

$$L(\varepsilon,\beta,\lambda) = \gamma \frac{1}{2}\|\varepsilon\| + \frac{1}{2}\|\beta\| + \lambda^{\mathrm{T}}[\mathbf{H}\beta - \mathbf{T} - \varepsilon] \tag{9}$$

where, $\lambda$ is the vector of $m$ Lagrange multipliers with the equality constraints of (8). The gradients of (9) with respect to the variables $\varepsilon$, $\beta$, and $\lambda$ are

$$\begin{cases} \dfrac{\partial L}{\partial \varepsilon} = \gamma \varepsilon^{\mathrm{T}} - \lambda \\ \dfrac{\partial L}{\partial \beta} = \beta^{\mathrm{T}} + \lambda \mathbf{H} \\ \dfrac{\partial L}{\partial \lambda} = \mathbf{H}\beta - \mathbf{T} - \varepsilon \end{cases}. \tag{10}$$

Setting the equations in (10) to zero, we get a system of linear equations with solution

$$\beta = \left(\frac{I}{\gamma} + \mathbf{H}^{\mathrm{T}}\mathbf{H}\right)^{\dagger} \mathbf{H}^{\mathrm{T}}\mathbf{T}, \tag{11}$$

which still can be solved using the Moore-Penrose pseudoinverse method. This regularization is also called ridge/Tikhonov regularization.

## 3. Proposed Techniques for Reducing Computational Cost of ELM at Test Time

In this section, we present our techniques for reducing the computational cost of the ELM at test time. We first provide empirical evidence that the input weights of an ELM can also be formed by values drawn randomly from the uniform discrete distribution on the set $\{-1,0,1\}$. Subsequently, we prove that the classification accuracy of both raw signals (integer) and normalized signals ($\ell_2$ norm set to $1$) are the same. Finally, we show that a matrix $\beta^{int}$ containing integer output weights can be constructed from a trained $\beta$ such that the classification accuracy of both are almost the same.

All tests in this section were performed using one small dataset called *bark* versus *woodgrain*, built from the Brodatz textures [18], and one large dataset called MNIST. We describe these datasets in Section 4.1. For all simulations we ran in this research, the bias values $b$ were fixed to zero as it resulted in better accuracies using our techniques. Moreover, we used the rectifier activation function $g(x) = \max(0, x)$ and set the regularization factor $\gamma = 1$.

### 3.1. Build the Input Weights Matrix Using Uniform Random Values Drawn from $\{-1,0,1\}$

As aforementioned in Section 2.2, the input weights matrix $\mathbf{W}$ can be drawn randomly from any continuous probability distribution. In this section we show empirical evidence that $\mathbf{W}$ can also be constructed using independent and identically distributed random variables drawn from the set $\{-1,0,1\}$.

**Empirical evidence 1.** *Let $\mathbf{W}$ be the input weights containing random real values drawn from the standard uniform distribution on the open interval $(0,1)$. Also, let $\mathbf{W}^{bin}$ be the input weights containing integers from the uniform discrete distribution on the set $\{-1,0,1\}$. The classification accuracies of the ELM built using $\mathbf{W}$ and the ELM built using $\mathbf{W}^{bin}$ are similar.*

The simulations that backed this finding were as follows. For each dataset, we created one hundred random $\mathbf{W}$ and $\mathbf{W}^{bin}$ and evaluated them on the test set. The averages of the classification accuracies are shown in Table 1. As one can note, the averages differ slightly for the classification of the small dataset *bark* versus *woodgrain*. Nevertheless, the averages for the classification of the large dataset MNIST is the same.

**Table 1:** Comparison of the classification accuracy on the test set between the originals $\mathbf{W}$ and the proposed binary versions $\mathbf{W}^{bin}$. These results are the average of the accuracy obtained by the classification using 50 different pairs of $\mathbf{W}$ and $\mathbf{W}^{bin}$. For both datasets, the number of hidden neurons were fixed at 2000 atoms. The datasets are described in Section 4.1.

| Distribution | *bark* versus *woodgrain* Accuracy % | MNIST Accuracy % |
|---|---|---|
| Continuous | 92.88 (1.13) | 95.96 (0.13) |
| Binary {-1,1} | 91.07 (1.23) | 95.96 (0.12) |

The use of a discrete distribution for building $\mathbf{W}$ is not new and has been done in [5]. Nevertheless, their approach uses random values drawn from set $\{-1/\sqrt{L}, 1/\sqrt{L}\}$ to build $\mathbf{W}$, which requires the use of floating-point multiplications. On the other hand, our approach is to drawn random uniform values from the set $\{-1, 0, 1\}$, which has the clear advantage of dismissing multiplications.

Therefore, it is feasible to trade each power demanding multiplication by a single low power addition when computing the most expensive step of the classification at test time, which is the matrix-vector multiplication between the input signal and the input weights. If the test signal being evaluated contains only integer values, the computational cost of the classification at test time can be even further reduced, since the addition of integers is one of the cheapest arithmetic operations performed in hardware.

### 3.2. Classify Integer Signals at Test Time

In this section, we prove that the classification accuracy of both raw signals (integer) and normalized signals ($\ell_2$ norm set to $1$) are the same. We prove to the case where the activation function of the SFLN is the rectifier activation function $g(x) = \max(0, x)$, also called hinge activation function, which sets all negative values to zero. Our choice for the hinge activation function is based on its good results in deep architectures [6, 12, 11, 20], besides it has also shown good results in the preliminary experiments we ran. Moreover, we set all biases $b$ to zero, as it has also been empirically shown better results.

**Theorem 1.** *Let $\beta$ be the output weight matrix trained with the normalized training set ($\ell_2$ norm set to $1$). The classification of both raw signals (integer values) and normalized signals ($\ell_2$ norm set to $1$) are exactly the same.*

*Proof.* Let $\mathbf{x}_{int}$ and $\mathbf{x}$ be respectively a raw vector from the test set and its normalized version, with $\|\mathbf{x}\| = 1$. Also, let $\mathbf{h} = g(\mathbf{W}^T \mathbf{x})$ be the values at the output of the hidden layer of the SFLN, which we call feature vector. Therefore,

$$\begin{aligned}
\mathbf{h} &= \max(0, \mathbf{W}^T \mathbf{x}_{int}) \\
&= g(\mathbf{W}^T \mathbf{x}) \\
&= g\left(\mathbf{W}^T \frac{\mathbf{x}_{int}}{\|\mathbf{x}_{int}\|}\right) \\
&= \max\left(0, \mathbf{W}^T \frac{\mathbf{x}_{int}}{\|\mathbf{x}_{int}\|}\right) \\
&= \max\left(0, \frac{1}{\|\mathbf{x}_{int}\|} \mathbf{W}^T \mathbf{x}_{int}\right).
\end{aligned}$$

As $\beta$ has been previously trained, the output vector $\mathbf{o}$ is computed as

$$\mathbf{o} = \max\left(0, \frac{1}{\|\mathbf{x}_{int}\|} \mathbf{W}^T \mathbf{x}_{int}\right) \boldsymbol{\beta}$$

$$= \max\left(0, \frac{1}{\|\mathbf{x}_{int}\|} \mathbf{W}^T \mathbf{x}_{int}\right) \boldsymbol{\beta}.$$

As the $\ell_2$ norm of any real vector different from the null vector is always greater than 0, then $\frac{1}{\|\mathbf{x}_{int}\|} > 0$, and, thus it can be put outside of the $\max(\cdot)$ operator without affecting $\mathbf{o}$

$$\mathbf{o} = \frac{1}{\|\mathbf{x}_{int}\|} \max\left(0, \mathbf{W}^T \mathbf{x}_{int}\right) \boldsymbol{\beta}.$$

The predicted class of $\mathbf{x}_{int}$ is

$$i^* = \underset{i}{\operatorname{argmax}}\{o_i\}\ i = 1, 2, \ldots, m,$$

which is the same as the class of $\mathbf{x}$ because it is not changed by term $\frac{1}{\|\mathbf{x}_{int}\|}$, as it only scales down all $o_i$. Hence, the classification of the signal in its raw format and its normalized version are equivalent.

Therefore, the input signals do not need to be normalized before classification. This technique when used together with the technique presented in the Section 3.2, reduces the cost of the matrix-vector multiplication between the input signal and the input weights by exchanging each expensive floating-point multiplication by a single low power integer addition. If one uses integer output weights, the whole classification at test time is performed using integer operations and, thus, the computation cost of the most expensive part of the classification at test time, which is the matrix-vector multiplication between the input signal and the input weights, it is feasible to trade each power demanding multiplication by a single low power addition. If the test signal being evaluated contains only integer values, the computational cost of the classification at test time can be even further reduced, since the addition of integers is one of the cheapest operations performed in hardware.

### 3.3. Create Integer Version of $\boldsymbol{\beta}$

In this section we present how to construct an integer version of the output weights trained using ELM. We also show empirical evidence that both original output weights and its integer version results in similar classification accuracy.

**Hypothesis 1.** *Let $(\mathbf{X}, \mathbf{T})$ and $(\mathbf{X}_t, \mathbf{T}_t)$ be respectively a training and test set. Also, let $\boldsymbol{\beta}$ be the matrix of floating-point output weights computed for $(\mathbf{X}, \mathbf{T})$ using (11). A matrix $\boldsymbol{\beta}^{int}$ containing the integer approximation of $\boldsymbol{\beta}$ can be built from $\boldsymbol{\beta}$ such that both $\boldsymbol{\beta}$ and $\boldsymbol{\beta}^{int}$ achieves similar classification accuracies on the test set $(\mathbf{X}_t, \mathbf{T}_t)$.*

An integer approximation of $\boldsymbol{\beta}$ can be computed as

$$\beta^{int} = \text{round}(\beta/\tau), \qquad (12)$$

were $\tau$ is the minimum absolute value of $\beta$ and $\text{round}(\cdot)$ is the function that rounds a floating-point value to its closest integer value. As according to (1) and (??), the predicted class of any instance $\mathbf{x} \in \mathbf{X}_t$ using $b\beta$, which is $\beta$ and its multiples, is the same for all $b \in \mathbf{R}$, it is expected that the classification accuracy of $\beta^{int}$ is similar to the classification accuracy of $\beta/\tau$.

After computing an integer version of $\beta$ it is very likely that the elements of $\beta^{int}$ are spread over a wide range of values and, thus, requiring a high number of bits to represent each element of $\beta^{int}$. We hypothesized that it is possible to reduce the bit precision of $\beta^{int}$ up to a point with no substantial decrease of its classification accuracy. To test our hypothesis we performed a simulation to check how the reduction of the bit precision of $\beta^{int}$ affects its classification accuracy. For this, we created many different versions of $\beta^{int}$ with reduced bit precision with each version $\beta^{int}_{(i+1)}$ built from $\beta^{int}_{(i)}$ by dividing its elements by 2 and rounding them to their nearest integer. The stop criterion is when the maximum element of $\beta^{int}$ is equal to 1. Next, we evaluate each $\beta^{int}_{(i)}$ on the test set. We also compute the bit precision of each $\beta^{int}_{(i)}$, which is the minimum number of bits required to store each of its elements. In order to get a better estimate of the accuracy versus the bit precision of $\beta^{int}$, we averaged the results from the simulation of 10 different classifiers per dataset. We show on Figure 1 the results of this simulation.

As shown on Figure 1, the classification accuracy of the $\beta^{int}$ on the *bark* versus *woodgrain* dataset does not change even when using almost half of the original precision. As for the MNIST dataset, one can use less than half of the original precision with no decrease of the classification accuracy whatsoever. These results are the average of 10 different classifiers $\beta^{int}$ built using $1000$ hidden nodes for the *bark* versus *woodgrain* dataset and $4000$ hidden node for the MNIST dataset.

## 4. Simulations

In this section, we evaluate how our techniques affect the accuracy of ELM on the datasets described in Section 4.1, where we also present the parameters we chose to evaluate our techniques as well as the steps we used to select the best models. At last, the analysis of the results from the best models we obtained comes in Section 4.2.

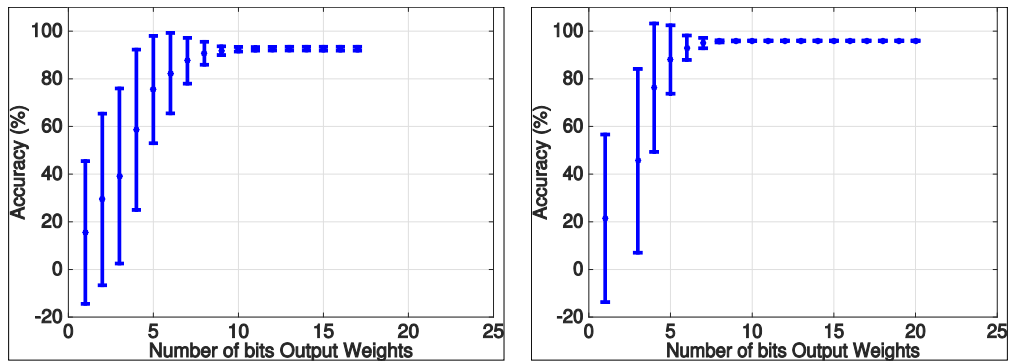

(**a**) *bark* versus *woodgrain*     (**b**) MNIST

**Figure 1:** Classification accuracy when the bit precision of the elements of the output weight matrix $\beta$ decreases. These results are the average of the accuracy obtained by the classification using 50 different pairs of $\mathbf{W}$ and $\mathbf{W}^{bin}$. For both datasets, the number of hidden neurons were fixed at 2000 atoms. The datasets are described in Section 4.1.

## 4.1 Datasets, Parameters, and Model Selection

The datasets we used to validate our techniques are:

- Brodatz textures *bark* versus *woodgrain* and *pigskin* versus *pressedcl*: The first two datasets contain patches of textures extracted from the Brodatz dataset [18]. As in [4], the first task consisted in discriminating between the images *bark* versus *woodgrain* and the second task consisted in the discrimination of *pigskin* versus *pressedcl*. First, we separated both images in two disjoint pieces and took the training patches from one piece and the test patches from the other one. As in [4], the training and test sets were built with 500 patches of the textures with size of $12 \times 12$ pixels. These patches were transformed into vectors and then normalized to have $\ell_2$ norm set to $1$.

- CIFAR-10 *deer* versus *horse*: The third binary dataset was built using a subset of the CIFAR-10 image dataset [9]. This dataset contains 10 classes of 60,000 $32 \times 32$ RGB images, with 50,000 images in the training set and 10,000 in the test set. Each image has 3 color channels and it is stored in a vector of $32 \times 32 \times 3 = 3072$ positions. The chosen images are those labeled as *deer* and *horse*.

- MNIST: The first multiclass dataset was the multiclass MNIST dataset [10], which contains 70,000 images of handwritten digits of size $28 \times 28$ distributed in 60,000 images in the training set and 10,000 images in the test set. As in [4], all images were preprocessed to have zero-mean and $\ell_2$ norm set to $1$.

- CIFAR-10 all classes: The last task consisted in the classification of all 10 classes from the CIFAR-10 image dataset.

As aforementioned in Section 3, the bias values $b$ were fixed at zero, the regularization factor fixed at $\gamma = 1$, and we used the rectifier activation function $g(x) = \max(0, x)$. Moreover, we varied the number of atoms on $L = \{10, 15, 25, 40, 60, 100, 160, 250, 400, 600, 1000, 1600, 2500, 4000, 6000\}$. These values were chosen as they are almost equidistant in $\log_{10}$ base, making the analysis visually easier.

For each dataset, the model selection process is as follow. For each number of hidden neurons $L$ we trained 96 original models $(\mathbf{W}, \beta)$ and 96 proposed models $(\mathbf{W}^{bin}, \beta^{int})$ using $80\%$ of the training set. With the remaining $20\%$ of the training set, we selected the best original and proposed models to evaluate the test set and the results we present in Section 4.2. We used the same steps to select both best original and proposed models, which are:

(i) Let $M$ be the set of models trained, $R = M(\mathbf{X})$ be the set of the classification accuracies on the reserved $20\%$ of the training set $\mathbf{X}$ using the models $M$, and $best = \max(R)$ be the best training accuracy from $R$.

(ii) From $M$, we create the subset $M_{best}$ that contains the models with results $R_{best} = R[\text{accuracy} >= 0.95 best]$.

(iii) Finally, we choose the model $(\mathbf{W}^*, \beta^*) \in M_{best}$ such that the $\beta^*$ has the lowest energy among all $(\mathbf{W}, \beta) \in M_{best}$.

**4.2. Results and Analysis**

In this section, the *original* results are the ones from the classification of the test set using the best original model built with the original ELM algorithm. Conversely, the *proposed* results are the ones obtained from the classification of the test set using the best proposed model built using our techniques for each dataset. The best original and proposed models are the ones selected using the methodology presented in Section 4.1.

We show the results of our simulations on the binary tasks in Figure 2. As shown in Figures 2(a), 2(b), and 2(c), our techniques incur in a limited reduction of the original classification accuracy. This can be better seen in Figures 2(d), 2(e), and 2(f), as they show the difference between the mean accuracy of the original and proposed for each number of neurons.

Figure 3 contains the results of the simulations on the tasks MNIST and CIFAR-10. Note that the results for both original and proposed models are quite similar, in contrast with the results for the smaller datasets shown on Figure 2. This is probably due to the larger number of instances in the larger datasets. This particular effect will be studied in a future investigation.

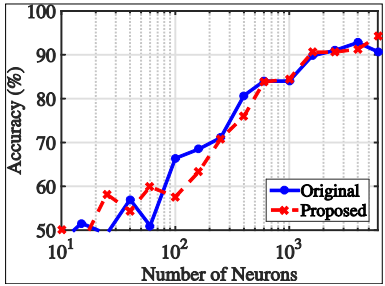
(**a**) *bark* versus *woodgrain*

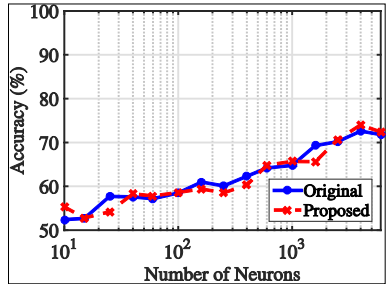
(**b**) *pigskin* versus *pressedcl*

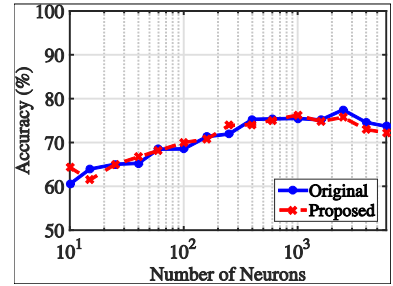
(**c**) CIFAR-10 *deer* versus *horse*

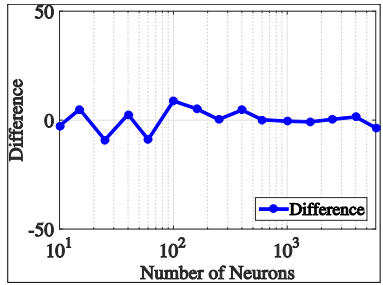
(**d**) *bark* versus *woodgrain*

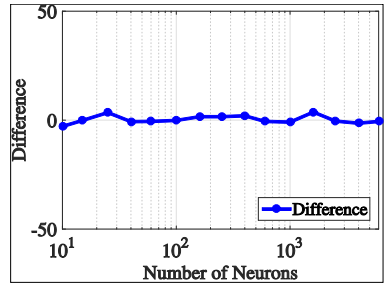
(**e**) *pigskin* versus *pressedcl*

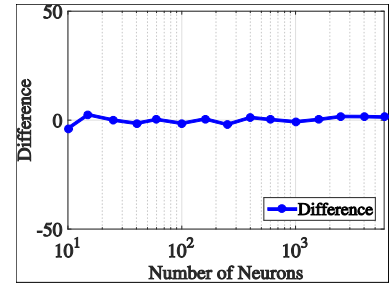
(**f**) CIFAR-10 *deer* versus *horse*

**Figure 2:** Comparison of classification accuracy on small datasets between the original ELM algorithm and the modified ELM with our proposed techniques. The datasets are described in Section 4.1.

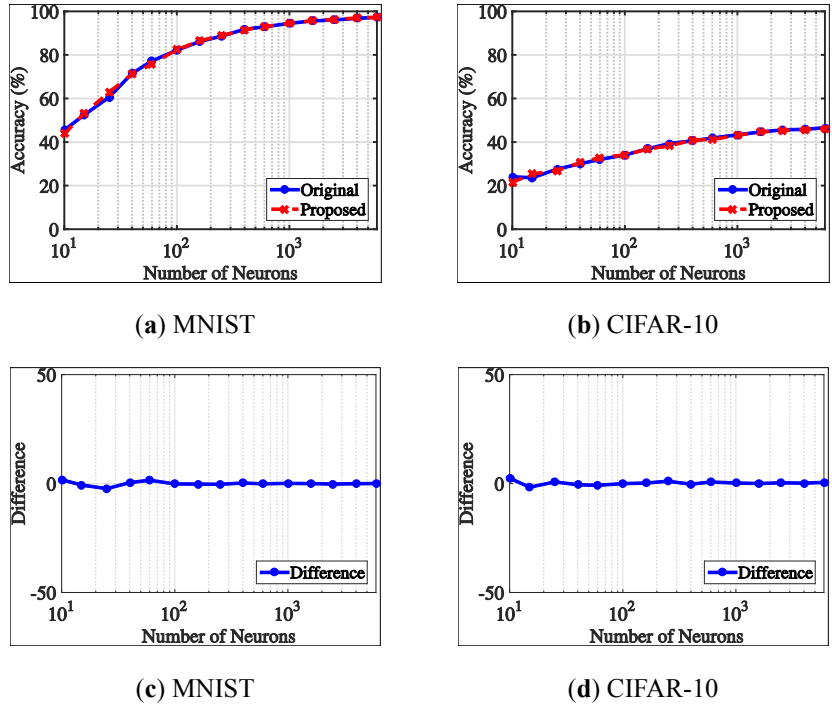

**Figure 3:** Comparison of classification accuracy on large datasets between the original ELM algorithm and the modified ELM with our proposed techniques. The datasets are described in Section 4.1.

The results we presented in this section indicate the feasibility of using low power integer operations in place of floating-point operations with no substantial difference in the classification accuracy. Moreover, our approach dismiss the use of multiplications in the feature mapping stage of the classification at test time, which is the part of the classification at test time that demands the most computational power. These substitutions we propose can significantly reduce the computational cost of classification at test time in FPGAs, which is important in embedded applications and in data centers of large corporations, where power consumption is critical.

## 5. Conclusion

This paper presented a set of techniques for computational cost reduction of test time operations of network classifiers based on extreme learning machine (ELM). Basically the techniques are: use random input weights drawn uniformly from the set $\{-1, 0, 1\}$ and, thus, eliminate all related multiplications; use at test time signals in its original format, i.d. non-normalized, and approximate the trained output weights to their respective closest integers, hence eliminating all floating-point operations.

We ran simulations using image datasets commonly used in the literature to assess classification algorithms and our results indicate it is feasible to dismiss the use of floating-point operations and use low power integer operations instead. Moreover, our techniques allow to trade each multiplication by a single addition in the most expensive part of the classification at test time, which is the random projection of test signals using the input weights. Altogether, our approach is specially important in applications running in FPGA, as it reduces the energy consumption when only integer operations are used.

## 6. Acknowledgments
This work was partially conducted during a scholarship financed by the Brazilian Federal Agency for Support and Evaluation of Graduate Education (Portuguese acronym CAPES) within the Ministry of Education of Brazil. We thank the Department of Electrical and Computer Engineering of the University of Texas at El Paso for allowing us access to the NSF-supported cluster (NSF CNS-0709438) used in all the simulations here described and also Mr. N. Gumataotao for his assistance with this cluster.

## References

[1] P. L. Bartlett. The sample complexity of pattern classification with neural networks: the size of the weights is more important than the size of the network. *IEEE Transactions on Information Theory*, 44(2), 1998.

[2] B. Cope, P. Y. Cheung, W. Luk, and L. Howes. Performance comparison of graphics processors to reconfigurable logic: a case study. *Computers, IEEE Transactions on*, 59(4):433–448, 2010.

[3] W. D. W. Deng, Q. Z. Q. Zheng, and L. C. L. Chen. Regularized Extreme Learning Machine. *Audio, Transactions of the IRE Professional Group on*, pages 389–395, Jan. 2009.

[4] A. Fawzi, M. Davies, and P. Frossard. Dictionary Learning for Fast Classification Based on Soft-thresholding. *International Journal of Computer Vision*, pages 1–16, Nov. 2014.

[5] P. Gastaldo, R. Zunino, E. Cambria, and S. Decherchi. Combining ELM with random projections. *IEEE Intelligent Systems*, 28(6):46–48, 2013.

[6] X. Glorot, A. Bordes, and Y. Bengio. Deep Sparse Rectifier Neural Networks. In *Proceedings of the 14th International Conference on Artificial Intelligence and Statistics*, pages 315–323, 2011.

[7] G.-B. Huang, Q.-Y. Zhu, and C.-K. Siew. Extreme learning machine: theory and applications. *Neurocomputing*, 70(1):489–501, 2006.

[8] A. Krizhevsky, I. Sutskever, and G. E. Hinton. ImageNet Classification with Deep Convolutional Neural Networks. *Advances in neural information processing systems*, pages 1097–1105, 2012.

[9] A. Krizhevsky. *Learning multiple layers of features from tiny images*. Computer Science Department, University of Toronto, 2009.

[10] Y. LeCun, L. Bottou, Y. Bengio, and P. Haffner. Gradient-based learning applied to document recognition. *Proceedings of the IEEE. Institute of Electrical and Electronics Engineers*, 86(11):2278–2324, 1998.

[11] A. L. Maas, A. Y. Hannun, and A. Y. Ng. Rectifier nonlinearities improve neural network acoustic models. In *Proc. ICML*, 2013.

[12] V. Nair and G. E. Hinton. Rectified linear units improve restricted boltzmann machines. In *Proceedings of the 27th International Conference on Machine Learning (ICML-10)*, pages 807–814, 2010.

[13] S. Ravishankar and Y. Bresler. Learning Sparsifying Transforms. *IEEE Transactions on Signal Processing*, 61(5):1072–1086, Jan. 2013.

[14] J. Schmidhuber. Deep learning in neural networks: An overview. *Neural networks : the official journal of the International Neural Network Society*, 61:85–117, Jan. 2015.

[15] S. Shekhar, V. M. Patel, and R. Chellappa. Analysis sparse coding models for image-based classification. In *IEEE International Conference on Image Processing. Proceedings*, 2014.

[16] S. W. Smith. *The scientist and engineer's guide to digital signal processing*. California Technical Pub. San Diego, Dec. 1997.

[17] J. Tang, C. Deng, and G.-B. Huang. Extreme Learning Machine for Multilayer Perceptron. *IEEE Transactions on Neural Networks and Learning Systems*, pages 1–1, May 2015.

[18] K. Valkealahti and E. Oja. Reduced multidimensional co-occurrence histograms in texture classification. *IEEE Transactions on Pattern Analysis and Machine Intelligence*, 20(1):90–94, 1998.

[19] V. N. Vapnik and V. Vapnik. *Statistical learning theory*, volume 1. Wiley New York, 1998.

[20] M. D. Zeiler, M. Ranzato, R. Monga, M. Mao, K. Yang, Q. V. Le, P. Nguyen, A. Senior, V. Vanhoucke, J. Dean, and others. On rectified linear units for speech processing. In *Acoustics, Speech and Signal Processing (ICASSP), 2013 IEEE International Conference on*, pages 3517–3521. IEEE, 2013.